\documentclass[conference]{IEEEtran}

\usepackage[pdftex]{graphicx}
\usepackage{amsmath}
\usepackage{hyperref}

\usepackage{color,soul, colortbl}
\definecolor{lightgreen}{rgb}{.8,1,.85}
\definecolor{lightblue}{rgb}{.8,0.85,1}
\definecolor{lightred}{rgb}{1,0.8,0.85}

\newcommand{\A}[1]{#1}

\begin{document}

\title{{Self-supervised}
Body Image Acquisition Using a Deep Neural Network for Sensorimotor Prediction}

\author{
\IEEEauthorblockN{Alban Laflaqui\`ere}
\IEEEauthorblockA{AI Lab\\
SoftBank Robotics Europe\\
Paris, France\\
alaflaquiere@softbankrobotics.com}
\and
\IEEEauthorblockN{Verena V. Hafner}
\IEEEauthorblockA{Adaptive Systems Group\\
Humboldt-Universit\"at zu Berlin\\
Berlin, Germany\\
hafner@informatik.hu-berlin.de}
}

\maketitle

\begin{abstract}
This work investigates how a naive agent can acquire its own body image in a {self-supervised} way, based on the predictability of its sensorimotor experience. Our working hypothesis is that, due to its temporal stability, an agent's body produces more consistent sensory experiences than the environment, which exhibits a greater variability. Given its motor experience, an agent can thus reliably predict what appearance its body should have. This intrinsic predictability can be used to automatically isolate the body image from the rest of the environment.
We propose a two-branches deconvolutional neural network to predict the visual sensory state associated with an input motor state, as well as the prediction error associated with this input. We train the network on a dataset of first-person images collected with a {simulated} Pepper robot, and show how the network outputs can be used to automatically isolate its visible arm from the rest of the environment. {Finally, the quality of the body image produced by the network is evaluated}.
\end{abstract}

\IEEEpeerreviewmaketitle

\section{Introduction}
\label{sec:introduction}

Going through different developmental phases, human children continuously acquire control over their own bodies \cite{Rochat2011}. Eventually, they can distinguish between self and others and predict the consequences of their own actions. 
This pre-reflective identification is often called the minimal self, and constitutes of two major components: {a} sense of agency, and a sense of body ownership \cite{gallagher2000philosophical}. 
These notions have recently become relevant also in developmental robotics. Firstly, computational models of the self can provide interesting insights into the processes of self-development in humans, and secondly an adaptive self-model can be crucial for intuitive and adaptive interaction in robotics.

{Different models have been proposed in the past few years to address one or more facets of this problem.}
In a recent {paper}, Lang, Schillaci and Hafner \cite{lang2018sense} presented a study on {the} sense of agency and object permanence in which a robot predicted its own arm movements using a convolutional neural network.
Hoffman et al. \cite{hoffmann2018robotic} investigated the formation of a body representation in a humanoid robot equipped with artificial skin during experiments on touch. Another study investigates the effects of self-touch in human infants related to their own motor actions, and suggests robotic models \cite{hoffmann2017development}.
An information-theoretic approach to the formation of body maps in robots has been proposed {in} \cite{KaplanH06}, {in which} the structure of the body map resulted from information distances between sensory data during a particular behaviour of the robot.
\begin{figure}[!t]
\centering
\includegraphics[width=0.9\linewidth]{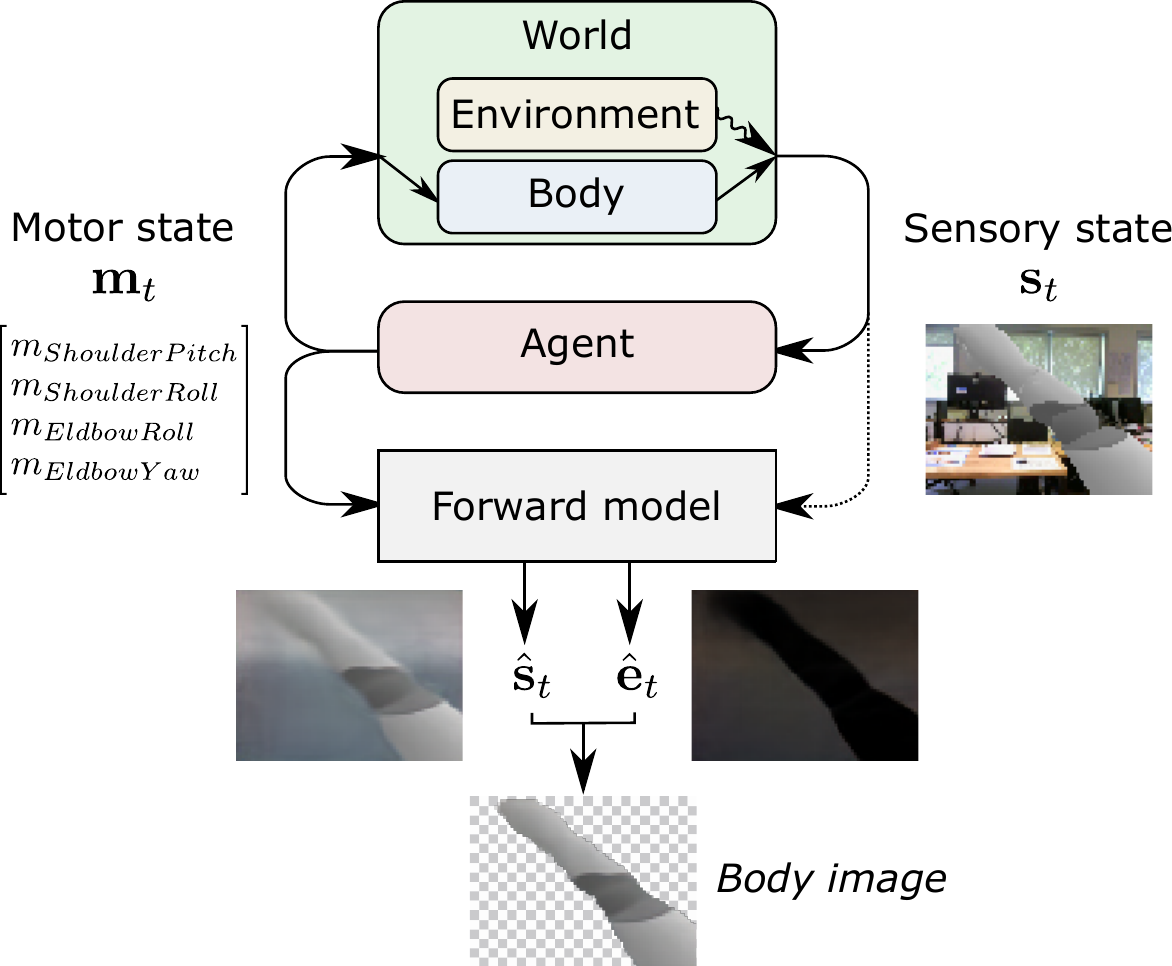}
\caption{Interaction between the agent and its environment, and forward model learned by the agent. Based on the predictability of the sensory inputs, the agent can isolate its body image from the rest of the environment.}
\label{fig:interaction}
\end{figure}
{An approach for body and non-body discrimination in robotics has been proposed by Yoshikawa et al.} \cite{Yoshikawa:2004}, {where they propose a method to approximate posture sensations by Gaussian distributions.}
{A predictive coding approach to generate visuo-proprioceptive patterns has been implemented by Hwang et al.} \cite{Hwang:2017} {on a simulated iCub robot. In this study, the robot was trained to imitate gestures of another robot or its mirror image displayed on a screen.}
{Hinz et al.} \cite{Hinz:2018} {present a study investigating prediction errors in tactile-visual data in both humans and robots. They use a rubber-hand illusion setup typically used to study properties of body ownership.}

\begin{figure*}[!t]
\centering
\includegraphics[width=0.9\linewidth]{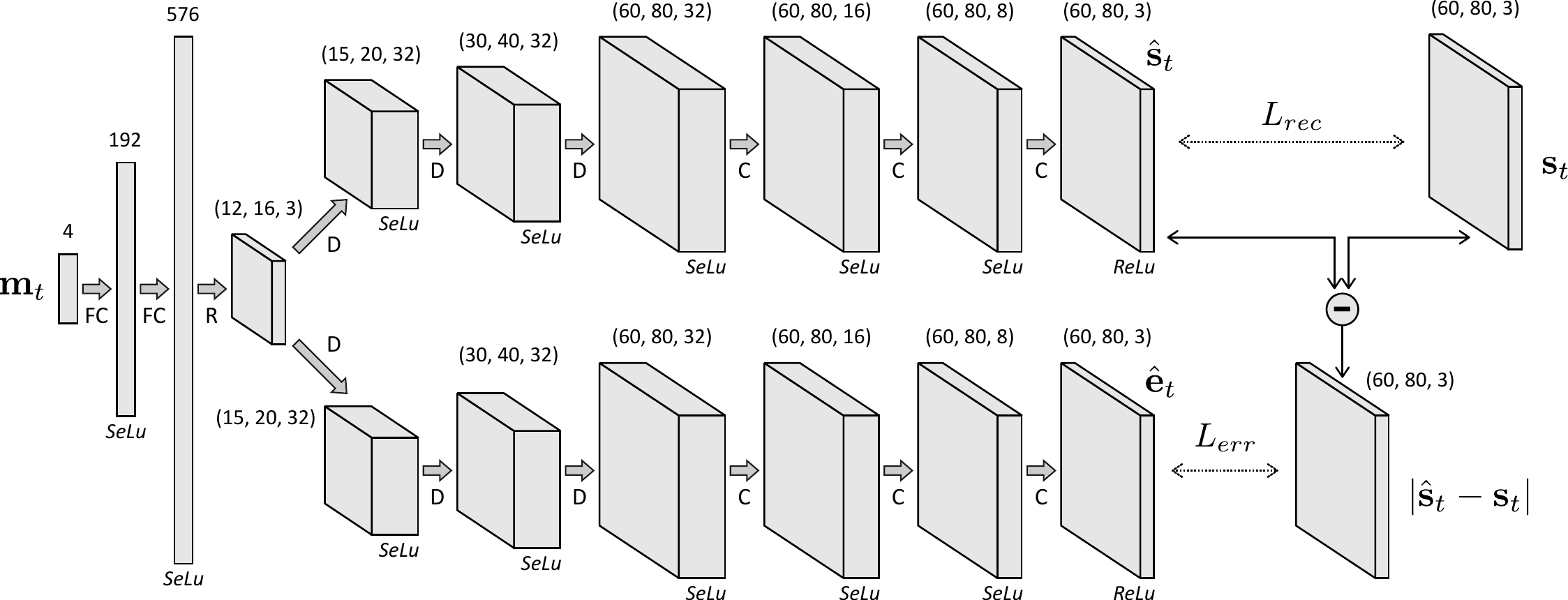}
\caption{The neural network architecture used to predict an output image $\hat{\mathbf{s}}_t$ and a prediction error $\hat{\mathbf{e}}_t$, given an input motor state $\mathbf{m}_t$. Two (de)convolutional branches are dedicated to produce these two respective outputs. In the diagram, FC stands for a fully-connected layer, R stands for a reshaping operation, D stands for a deconvolutional layer, and C stands for a convolutional layer.}
\label{fig:network}
\end{figure*}
 
Gallagher \cite{Gallagher:1985} argues that body image and body schema are separate concepts. He suggests that the body image is a conscious representation of the body, whereas the body schema is a prepersonal structural mapping.
In artificial systems, a clear distinction between these two concepts is more difficult to make{, and their respective definition often varies}.
In this paper, we use the term \emph{"body image"} {in a very literal sense,} as the "appearance of the body in the visual flow", and we argue that body image acquisition depends on the predictive capabilities of the agent.
Extending the studies of Lang et al. \cite{lang2018sense}, we suggest a mechanism for an artificial agent to acquire its body image from scratch through predictive processes in {self-supervised} way.

The paper is structured as follows: section \ref{sec:methodology} {presents the methodology and neural network used in the approach.}
Section \ref{sec:experiments} presents the {experimental setup.}
Section \ref{sec:results} presents {the qualitative and quantitative results}. Finally, we discuss the results and conclusions in section \ref{sec:conclusion}.

\section{Methodology}
\label{sec:methodology}

\subsection{Approach}
\label{sec:approach}

The objective of this work is to study how an agent could learn its own body image {(appearance)} in a {self-supervised} way.
Our working hypothesis is that {when predicting sensorimotor experiences, the body appears as a primary source of predictability, as it} is always present and does not change, or only at a very slow pace. As a consequence, when {reaching motor states}, the ``appearance" of the body in the sensory flow is consistent throughout the exploration of the environment. Compared to the rest of the sensory flow induced by the environment, which the agent does not directly control, this \emph{body image} should thus be significantly more predictable.

We can formalize this intuition by considering the mapping between the agent's motor state and sensory state.
We denote $\mathbf{m}_t = [m_{1,t}, \dots, m_{N_m,t}]$ the {proprioceptive} motor state, or posture, in which the agent is at time $t$, where each $m_{n,t}$ corresponds to the {static} configuration of a joint. Similarly, we denote $\mathbf{s}_t = [s_{1,t}, \dots, s_{N_s,t}]$ the sensory state that the agent receives from its exteroceptive sensors at time $t$, where each $s_{n,t}$ is an individual sensory component (see Fig.~\ref{fig:interaction}).
\\
We assume in this work that the sensors are such that they provide an instantaneous reading of the state of the world, and that, for any motor state $\mathbf{m}_t$, the sensory state $\mathbf{s}_t$ can be divided into two subsets: {$\mathcal{B}_{\mathbf{m}_t}$ and $\overline{\mathcal{B}_{\mathbf{m}_t}}$}.
The subset {$\mathcal{B}_{\mathbf{m}_t}$} gathers all components $s_{i,t}$ associated with the body, while its complement gathers {the ones} associated with the environment.
Said otherwise, we assume that the elementary sensory excitations $s_{i,t}$ are not due to a mixed contribution of the body and the environment. This is typically the case with a camera as, for a given body posture, a subset of pixels ({and their respective channels}) corresponds to body parts in the field of view, while other pixels correspond to the environment.

Without a priori knowledge about the state of the environment, the mapping $\mathbf{m}_t \to \mathbf{s}_t$ is not deterministic. Putting aside potential sensorimotor noise, the uncertainty about the state of the environment limits the ability to predict $\mathbf{s}_t$.
However, if the body appearance stays temporally consistent, we hypothesize that the subset {$\mathcal{B}_{\mathbf{m}_t}$} should exhibit a significantly lower \A{variability in time} than {$\overline{\mathcal{B}_{\mathbf{m}_t}}$}, i.e.:
\begin{equation}
\begin{aligned}
    & s_{i,t} \in \mathcal{B}_{\mathbf{m}_t} \Rightarrow \mathrm{Var}\big(s_{i,t} | \mathbf{m}_t\big) \longrightarrow 0,\\
    & s_{i,t} \in \overline{\mathcal{B}_{\mathbf{m}_t}} \Rightarrow \mathrm{Var}\big(s_{i,t} | \mathbf{m}_t\big) \longrightarrow \epsilon \gg 0,
\end{aligned}
\label{eq:variance}
\end{equation}
where $s_{i,t}$ is treated as a random variable {over time}.
In an environment which exhibits a sufficient amount of variability, this intrinsic difference can be used, in a data-driven way, to distinguish the sensory components which belong to the body image and to the environment (symmetrizing the implication arrows of Eq.\eqref{eq:variance}).
{Note that the typical RGB excitations of a pixel are here considered as separate components $s_{i,t}$.}

More formally, the agent can learn a forward model mapping from $\mathbf{m}_t$ to a prediction $\hat{\mathbf{s}}_t$ of $\mathbf{s}_t$, as well as to a prediction $\hat{\mathbf{e}}_t$ of the error $| \hat{\mathbf{s}}_t - \mathbf{s}_t |$ (see Fig.~\ref{fig:interaction}).
According to Eq.~\ref{eq:variance}, the elementary prediction errors $\hat{e}_{i,t}$ should be significantly lower if $s_{i,t} \in \mathcal{B}_t$ than if $s_{i,t} \in \overline{\mathcal{B}_t}$, allowing to distinguish the two subsets based on the accuracy of the forward model.

\begin{figure*}[!t]
\centering
\includegraphics[width=1\linewidth]{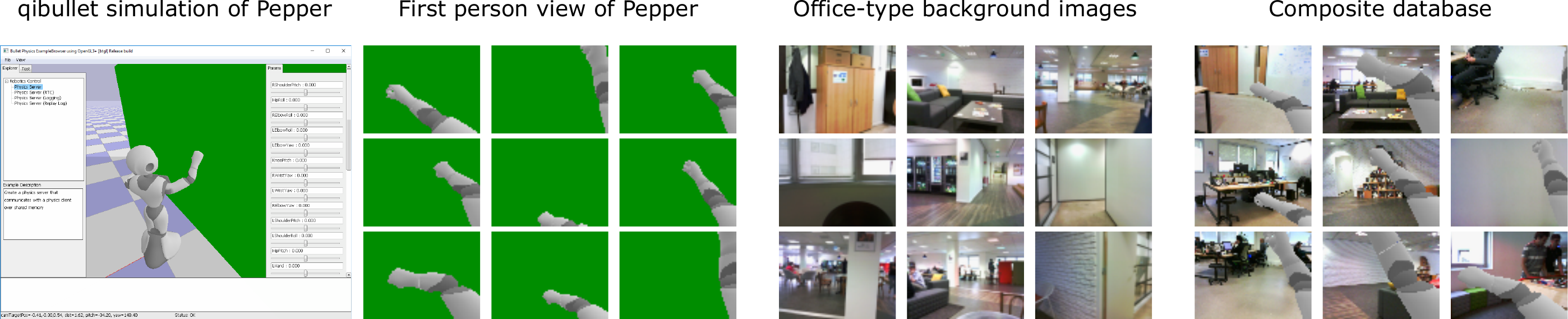}
\caption{The qibullet simulator is used to generate first person view images of the Pepper robot's right arm moving in front of a green background. This background is then replaced by office-type images collected by a real Pepper robot. The final training dataset consist in these composite images and their corresponding motor states (posture).}
\label{fig:dataset}
\end{figure*}

\subsection{Neural network architecture}
\label{sec:neural network architecture}

We use a deep neural network to learn both the forward mapping $\mathbf{m}_t \to \hat{\mathbf{s}}_t$ and the error prediction mapping $\mathbf{m}_t \to \hat{\mathbf{e}}_t$.
As displayed in Fig.~\ref{fig:network}, the network first passes the input motor state $\mathbf{m}_t$ through two fully-connected layers. The last of these layers is then reshaped to form a low resolution image-like representation that can be spatially processed for deconvolution.
This intermediary representation is then fed to two distinct branches which respectively predict the output image $\hat{\mathbf{s}}_t$ and the prediction error $\hat{\mathbf{e}}_t$. Each branch is composed of three successive deconvolutional layers and three convolutional layers. The three deconvolution layers upscale the low resolution representation to the final size of the image. Following the good practice proposed in \cite{odena2016deconvolution}, the deconvolution operation consists in an upscaling operation, followed by a convolution. The next three convolutional layers perform typical convolution operations \cite{lecun1990handwritten}, and progressively reduce the depth of the input from 32 channels to 3, corresponding to the RGB channels of the pixels.
{In each branch, all layers use the SeLu activation function} \cite{klambauer2017self} {---which performs an internal normalization of the neurons' activation---, except the last convolutional layer which uses the ReLu activation function} \cite{nair2010rectified}, {in order to guarantee the positivity of each $\hat{s}_{i,t}$ and $\hat{e}_{i,t}$}. Throughout the network, all convolutions are done using kernels of size $3 \times 3$, with a stride of 1.

A different loss is associated with each branch of the network. The first branch (top one in Fig.~\ref{fig:network}) {outputs a predicted image $\hat{\mathbf{s}}_t$, and its associated reconstruction loss is}:
\begin{equation}
    L_{rec} = \frac{1}{N N_s} \sum_{k=1}^N \sum_{i=1}^{N_s} | \hat{s}_{i,k} - s_{i,k} |,
\end{equation}
where $s_{i,k}$ is {a sensory component} for input $k$, and $|.|$ is the absolute operator.
It corresponds to the mean value of the $L^1$ norm between $\hat{\mathbf{s}}_t$ and $\mathbf{s}_t$, normalized by {the number of sensory components (three times the number of pixels).}
\\
The second branch (bottom one in Fig.~\ref{fig:network}) {outputs the predicted absolute prediction error $\hat{\mathbf{e}}_t$ between the predicted image $\hat{\mathbf{s}}_t$ and the ground-truth image $\mathbf{s}_t$, and its associated loss is}:
\begin{equation}
    L_{err} = \frac{1}{N N_s} \sum_{k=1}^N \sum_{i=1}^{N_s} \big| \hat{e}_{i, k} - |\hat{s}_{i,k} - s_{i,k}| \big|.
\end{equation}
It corresponds to the mean value of the $L^1$ norm between $\hat{\mathbf{e}}_t$ and the actual prediction error $| \hat{\mathbf{s}}_t - \mathbf{s}_t |$, normalized by {the number of sensory components.}
Finally, the total loss is a {weighted} composition of the these two losses:
\begin{equation}
    L = L_{rec} + \alpha . L_{err},
\end{equation}
{where $\alpha$ denotes a scalar relative weight.}\\
This total loss is minimized using the ADAM optimizer \cite{kingma2014adam}, for $5\times10^{4}$ iterations, and with a learning rate linearly decreasing from $10^{-3}$ to $10^{-5}$ during training. At each iteration a random mini-batch of {$N=100$} samples is fed to the network to compute the (stochastic) gradient.
{Finally, $\alpha$ is set to increase linearly from $0$ to $1$ during the first $2.5\times10^{4}$ iterations. It helps stabilizing the convergence by first focusing the optimization on the image prediction branch, so that its output can be used as a reliable target for the second branch.}

\section{Experiments}
\label{sec:experiments}

\subsection{Sensorimotor data generation}
\label{sec:sensorimotor data generation}

In order to test our approach, we create{d} a dataset of images in which a {simulated} Pepper robot \cite{pepper} observes its own right arm in different environments.
In order to quickly generate large datasets without directly using the physical robot and to easily assess a ground-truth body image mask, we created a {synthetic} experimental dataset by composing images.

First, the realistic Pepper simulator qibullet \cite{qibullet}, {was} used to quickly generate $8000$ right arm configurations $\mathbf{m}_t$. They {were} randomly generated by uniformly drawing the 4 following joint orientations \mbox{$m_{ShoulPitch} \in [-1, 1]$}, \mbox{$m_{ShoulRoll} \in [-1, 0]$}, \mbox{$m_{ElbRoll} \in [0, 1]$}, \mbox{$m_{ElbYaw} \in [-1, 1]$} (radians).
The motor exploration thus {resembles} a typical motor babbling \cite{meltzoff1997explaining}.
During this exploration, the simulated Pepper is placed in front of a green wall{, with the head oriented towards the right arm (downward pitch of $0.253$, rightward yaw of $0.7$ (radians))}.
For each {$\mathbf{m}_t$}, the $(240 \times 320)$ pixels image captured by the robot's camera is recorded.
As can be seen in Fig.~\ref{fig:dataset}, these images correspond to a first person view of the arm in front of a uniform green surface. This clear bimodal structure allows use to easily replace the green surface with any desired background.
We {filled it} with {random} images from a previous dataset collected {while a real Pepper robot moved} in an office-like environment.
The robot's body is not visible in this previous dataset, which means that the arm images generated using qibullet can be embedded in them without ambiguity.
The whole creation of the dataset is illustrated in Fig.~\ref{fig:dataset}, where one can see the arm configurations and the background images used to compose the experimental dataset.

\begin{figure*}[!t]
\centering
\includegraphics[width=1\linewidth]{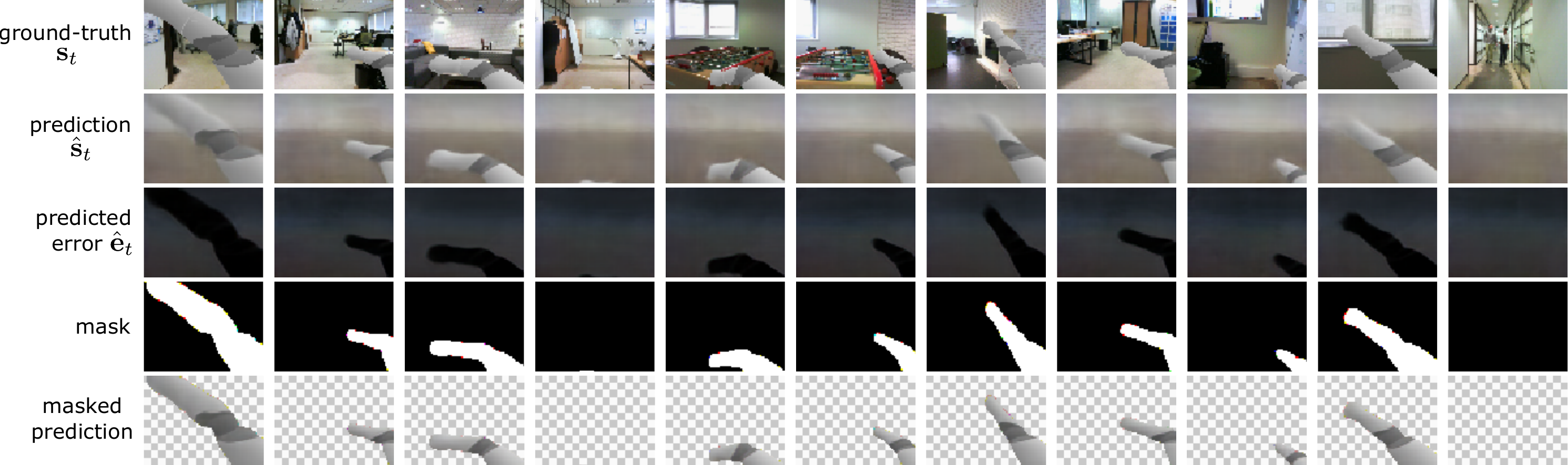}
\caption{Evaluation of the network outputs on a random subsampling of the training dataset. First row: ground-truth composite image from the dataset. Second row: image $\hat{\mathbf{s}}_t$ predicted by the network, given the corresponding motor state $\mathbf{m}_t$. Third row: prediction error $\hat{\mathbf{e}}_t$ predicted by the network, given the motor state. Fourth row: body mask automatically derived from $\hat{\mathbf{e}}_t$. Fifth row: predicted body image after applying the mask to $\hat{\mathbf{s}}_t$.
Note that each plot is displayed as a RGB image, including the mask, as the 3 channels are considered independently in the data processing.}
\label{fig:results}
\end{figure*}

Note that {in} this compositional approach, the ambient lighting of the scene (background) does not affect the appearance of the robot's arm. Due to its probabilistic nature, we however expect our approach to be robust to small changes {in} lighting conditions, such as the ones observed in the office-type background dataset.
{This however needs to be confirmed by future experiments.}

Before being fed to the network, both the motor states and sensory {states} (images) are pre-processed.
The motor states are normalized such that each component spans the $[-1, 1]$ interval, and the sensory {states} are normalized such that all {components} lie in $[0, 1]$ (instead of $[0, 255]$ originally). Finally, the input images are downsampled to a $(60 \times 80)$ resolution, in order to limit the network size and computation time (although no theoretical limitation prevents the approach from scaling to greater resolutions).

\subsection{Body image extraction}
\label{sec:body image extraction}

Our working hypothesis is that for each motor state $\mathbf{m}_t$ it should exist a subset of {sensory components} for which predictability is significantly higher than for {other ones}. Those {components} should correspond to the consistent part of the robot's visual experience: its body image.
We thus propose to {automatically} isolate this body image by looking at the predicted prediction error {$\hat{\mathbf{e}}_t | \mathbf{m}_t$}. If its distribution is indeed bimodal, we propose to set a threshold {$\mathcal{T}$} between these two {modes} and to consider any {component} with a predicted error {$\hat{e}_{i,t} | \mathbf{m}_t$} inferior to {$\mathcal{T}$} as part of the body.
This way we can create a binary mask to isolate the body image from the rest of the input image.\\

{The code used to generate the data, and train and test the network is available at:} \url{https://github.com/alaflaquiere/learn-masked-body-image}.

\section{Results}
\label{sec:results}

After training, we qualitatively and quantitatively evaluate the mappings learned by the neural network.
Figure~\ref{fig:results} shows the predicted image $\hat{\mathbf{s}}_t$ and the predicted prediction error $\hat{\mathbf{e}}_t$ for $11$ random samples from the training set.\\
Firstly, we can see that the predicted images contain a meaningful approximation of the appearance of the arm. Note that the network also predicts the absence of the arm when the motor state moves it out of the field of view (see the last column of Fig.~\ref{fig:results}).
The arm appears in front of a background made of two relatively uniform horizontal stripes. The brighter upper stripe seems to statistically correspond to walls and windows which tend to be white in the background dataset. The darker lower stripe seems to statistically correspond to the floor which tends to be darker in the dataset.
Apart from this statistical distinction, the background of the predicted images does not contain any specific pattern from the original ground-truth images. In order to minimize its prediction error, the network thus learned to output the expectation of each unpredictable background {sensory component.}
\\
Secondly, the predicted error displays a similar structure. The area of the image corresponding to the arm is associated with a very low predicted prediction error {(darker)}, while the background is associated with greater errors {(lighter)}.
It thus seems that our working hypothesis was correct: based on the predictability of the visual input sub-components, it is possible differentiate the body image from the rest of the environment in {a self-supervised} way.

\begin{figure}[!t]
\centering
\includegraphics[width=0.8\linewidth]{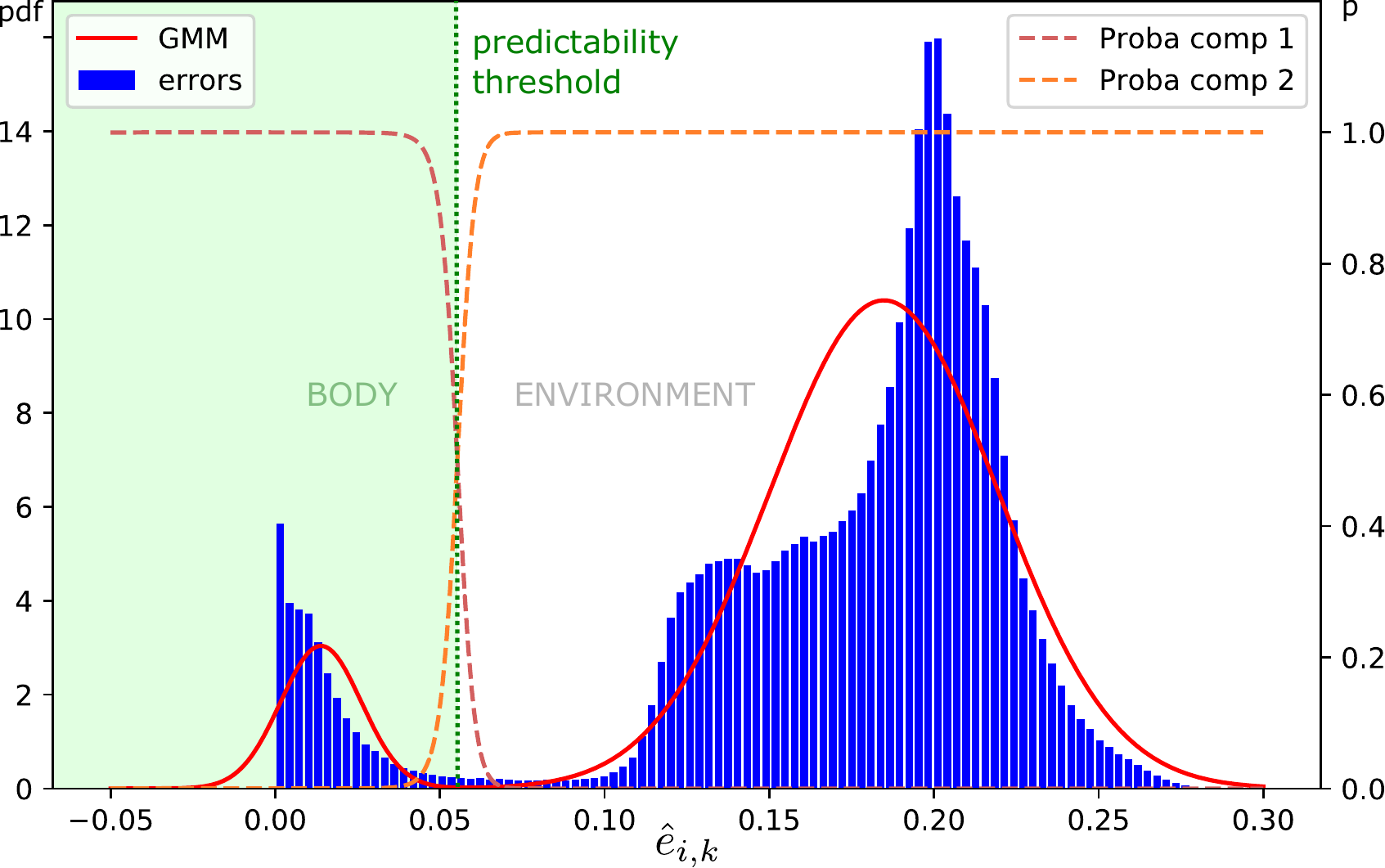}
\caption{Normalized histogram of all the predicted prediction errors $\hat{\mathbf{e}}_{i,k}$ over 100 random inputs from the dataset. The distribution appears to be bimodal, with a lower mode corresponding to highly predictable sensory components, and a higher mode corresponding to unpredictable ones. The threshold $\mathcal{T}=0.056$ is set at the intersection of the two modes.}
\label{fig:histogram}
\end{figure}

Figure~\ref{fig:histogram} shows a normalized histogram of all the predicted prediction errors $\hat{\mathbf{e}}_{i,k}$ produced by the network for $100$ random motor states from the training dataset.
{As expected,} the distribution appears to be bimodal.
We fit {it} with a \mbox{2-component} Gaussian Mixture Model and define a threshold at the intersection of the two components, i.e. {$\mathcal{T}=0.056$}. This threshold is used to automatically distinguish the {sensory components} $\hat{s}_{i,t}$ belonging to the body image {$\mathcal{B}_t$ ($\hat{e}_{i,t} \leq \mathcal{T}$)} from the ones belonging to the environment {$\overline{\mathcal{B}_t}$ ($\hat{e}_{i,t} > \mathcal{T}$)}.
It allows us to compute a mask {on the sensory components} to isolate the body image, as displayed in Fig.~\ref{fig:results}.
{Note that in this process, the R, G, and B channels of each pixel are treated independently, assuming minimal a priori knowledge about the structure of the sensory state. This potentially allows a mismatch between the different color components of the body image.}
{Finally, this mask of depth 3 is applied to the predicted image $\hat{\mathbf{s}}_t$ by making transparent the components not in the mask.}
The resulting body image, cleaned from the poorly {predictable} background, is displayed in Fig.~\ref{fig:results} as well.
\\
We can see that the appearance of the arm in the masked predicted images is very close to the one in the ground-truth images. The biggest disparities are located at the arm tip (hand), which appears to be the most difficult part to reconstruct. This can be explained by the fact that the appearance and localization of the hand in the image changes the most rapidly as a function of the joint configuration. On the contrary, the upper arm, closest to the shoulder and to the camera, is the most consistently reconstructed {part}, as its appearance changes the least as a function of the joint configuration.

\begin{figure}[!t]
\centering
\includegraphics[width=1\linewidth]{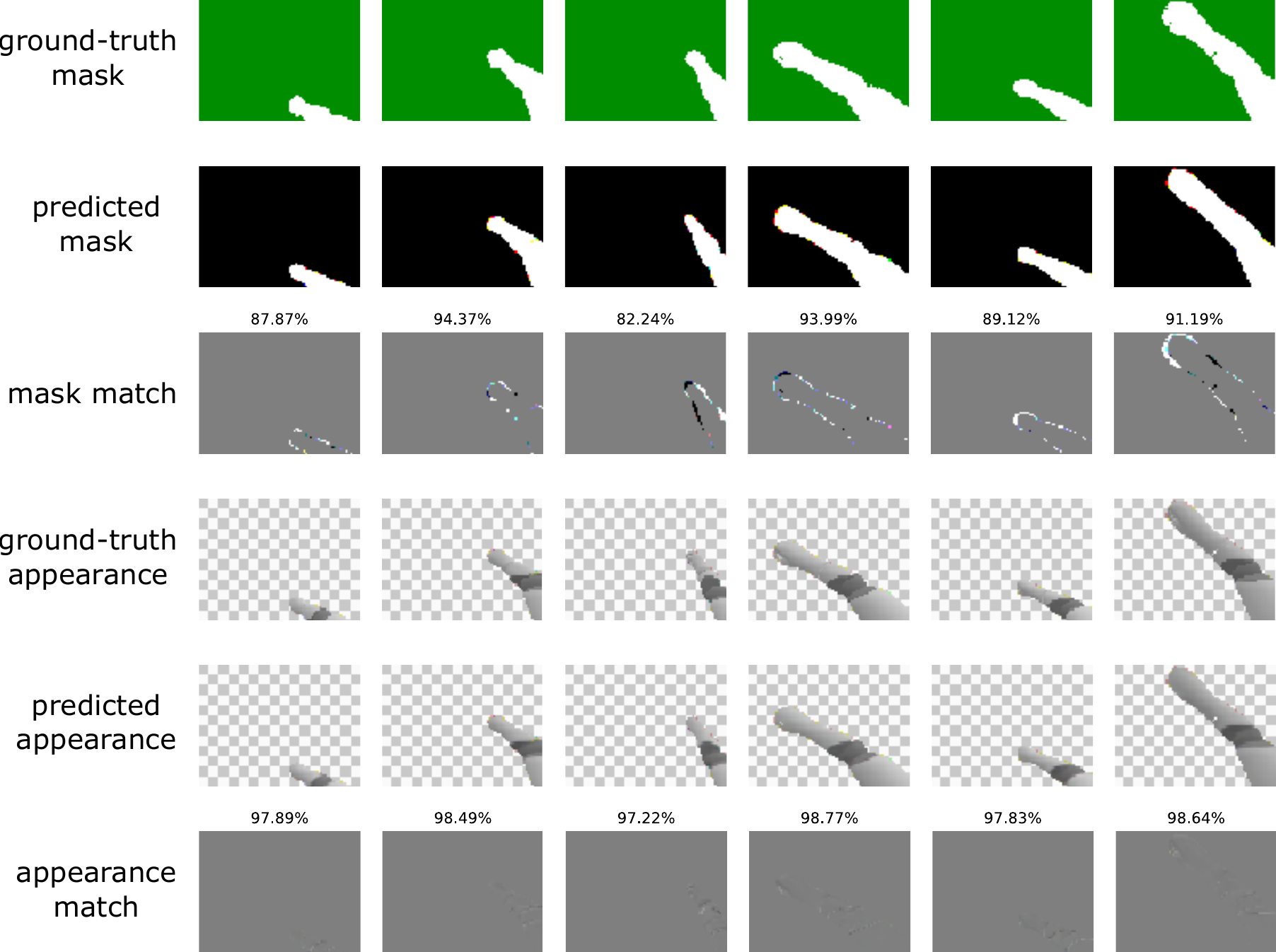}
\caption{Visualization of the match between the ground-truth body mask and the estimated body mask (first three rows), and of the match between the ground-truth appearance of the arm and the predicted appearance of the arm after application of the mask (last three rows).}
\label{fig:quantitative_evaluation}
\end{figure}

{We introduce two measures to quantify the quality of the learned body image.}
First, the \emph{mask match} corresponds to {the Intersection over Union (IoU) of the mask derived from the network's output and the ground-truth mask (non-green pixels in the image before composition).}
\begin{equation}
\text{{\it mask match}} = 
\#(\hat{\mathcal{B}}_{\mathbf{m}_t} \cap \mathcal{B}_{\mathbf{m}_t})
/
\# (\hat{\mathcal{B}}_{\mathbf{m}_t} \cup \mathcal{B}_{\mathbf{m}_t})
.
\end{equation}
Second, the \emph{appearance match} is defined as:
\begin{equation}
\text{{\it appearance match}} = 1 - \frac{1}{N_\mathcal{B}}
\sum_{i=1}^{N_\mathcal{B}}
\mathbf{1}_\mathcal{B} | \hat{s}_{i,t} - s_{i,t} |,
\end{equation}
{where $\mathbf{1}_\mathcal{B}$ is equal to $1$ if $s_{i,t} \in \hat{\mathcal{B}}_{\mathbf{m}_t} \cap \mathcal{B}_{\mathbf{m}_t}$ and $0$ otherwise, and $N_\mathcal{B}$ is the number of components in this mask intersection.}
{Note that in each measure, the three RGB channels of each pixel are considered independently.}
\\
Figure~\ref{fig:quantitative_evaluation} displays 6 visualizations of mask and appearance matches for random inputs of {a testing dataset generated the same way as the training dataset.}
We can see that most errors happen at the edge of the arm, where pixels values are the most likely to quickly change as a function of the motor joints, due to the environment being unpredictable.
We can also see that the errors in the arm {appearance} are insignificant, and barely visible as a consequence {(see last row)}.
Over the whole {testing dataset of 2000 samples}, the mask match is equal to $88.8 \pm 13.3 \%$, 
and the appearance match is equal to $98.3 \pm 0.7 \%$. 
{The match between the ground-truth body image and the one produced by the network is thus very good.}

\begin{figure}[!t]
\centering
\includegraphics[width=1\linewidth]{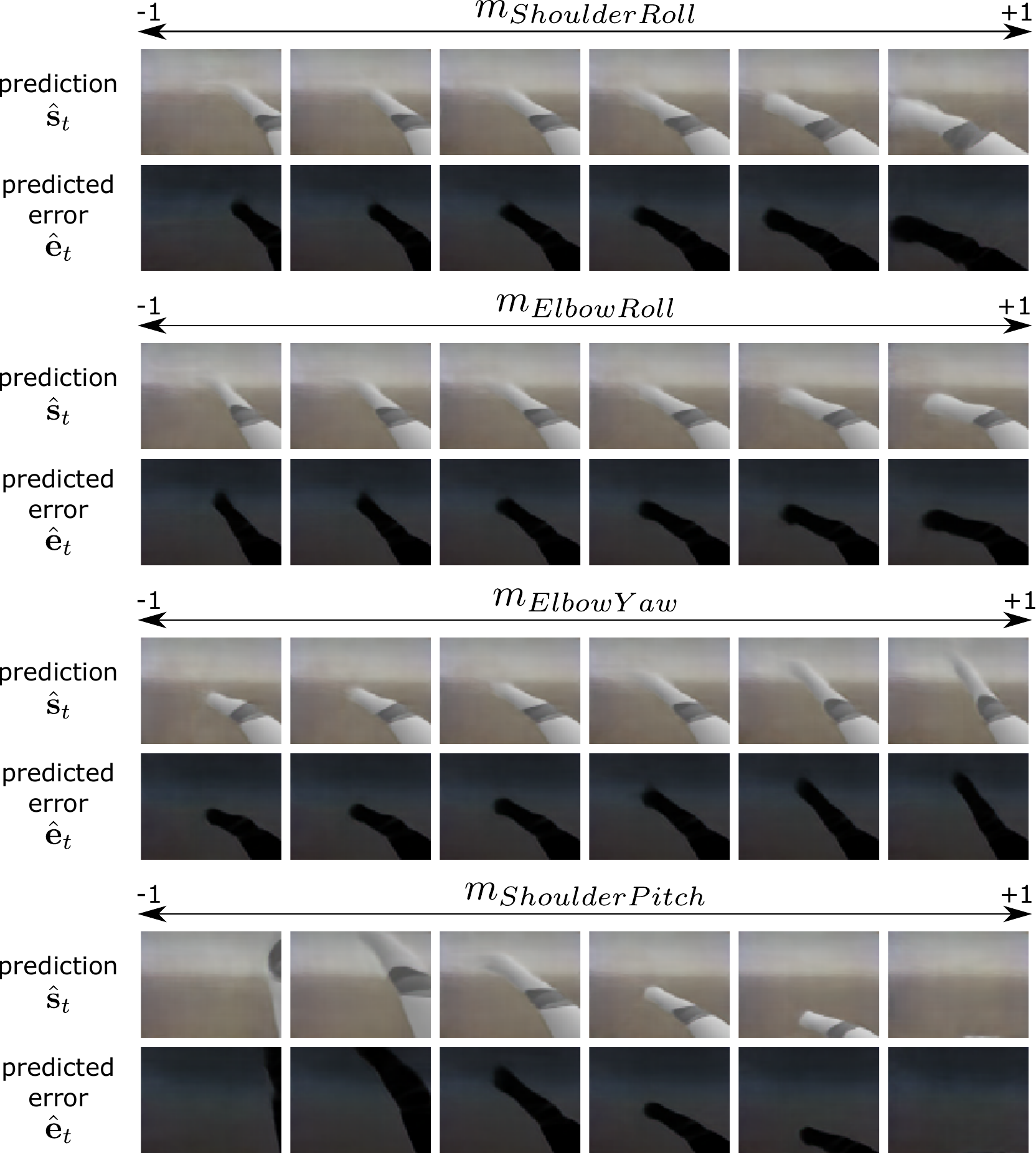}
\caption{Sampling of the motor space. Starting from the reference motor state $[0,0,0,0]$, each motor dimension is individually crossed from -1 to 1, and the corresponding network outputs are displayed.}
\label{fig:results2}
\end{figure}

Finally, the network can act as a (deterministic) conditional generative model.
We can sample the input motor space and let the network generate an output image and output prediction error.
Figure~\ref{fig:results2} shows the result of such a sampling when varying each motor component independently, starting from the reference arm configuration $[0,0,0,0]$.
We can see that the arm appearance and the associated low prediction error mask stay consistent throughout the motor space. The neural network {thus seems to generalize and to accurately predict the arm appearance over the whole motor space.}

\section{Conclusion}
\label{sec:conclusion}

We presented an algorithm and experiments for {autonomous} body image acquisition.
This work extends the studies of Lang et al. \cite{lang2018sense} by a mechanism {to distinguish sensory components that belong to the body image from those that belong to the environment, from scratch, in a self-supervised way.
It relies on the hypothesized intrinsic difference in variability of those two kinds of components, that a robot can capture when trying to predict its sensory state given an input motor state.}

{Like in the previous one, only movements of a single arm have been considered in this study. However this work could potentially be extended to more complex movements, including other limbs or the head itself.}
As shown by Schmerling et al. \cite{Schmerling2015}, the resulting change in the visual image caused by the head motion could {indeed} be included in the predictive learning process.
{The only theoretical limitation of such an extension is the number of samples required to correctly estimate the conditional statistics of the sensory experience as the dimension of the motor space increases.}

We have shown that the prediction of the visual information {associated with a motor state} is crucial for {the formation of the body image}.
From a predictive coding perspective, the body image would in this way correspond to the component of the sensory experience which is reliably and quickly predictable, on a developmental scale, {given the motor experience that the agent generates itself}.
{The importance of motor information in this formative process suggests a strong connection between the sense of agency, body ownership, body schema, and body image.}
{Another possible extension of this work would be to emphasize this motor aspect even more by introducing a dynamical system formulation of the problem,} like has been proposed in \cite{lang2018sense}. {A network could for instance be optimized to predict} the future sensory state $\mathbf{s}_{t+1}$, given the current sensory state $\mathbf{s}_{t}$ and a motor command $\mathbf{m}_{t}${, instead of a static posture like in the current formulation}.
{It would also be interesting to investigate if it is possible for an extended version of the network to simultaneously learn to infer its current body posture $\mathbf{m}_t$, using an auto-encoder paradigm, and to isolate its body image, based on the latent code learned by this auto-encoder.}

{Finally, it is important to note again that we used in this work the term ``body image'' in a literal sense to refer to the \emph{appearance} that the body has in a visual flow. The same term is often used to refer to more complex notions covering different psychological concepts, and whose definition can vary. Many open questions remain regarding the complete modeling of the notion of body image, and its coupling with other notions like body schema, body ownership, or the sense of agency.}

\bibliographystyle{IEEEtran}
\bibliography{biblioICDL2019}

\end{document}